\documentclass[10pt,twocolumn,letterpaper]{article}

\usepackage{iccv}
\usepackage{times}
\usepackage{epsfig}
\usepackage{graphicx}
\usepackage{amsmath}
\usepackage{amssymb}
\usepackage{bbm}
\usepackage{dsfont}
\usepackage{multirow}
\usepackage{enumitem}
\usepackage{pifont}
\usepackage{comment}
\usepackage{caption}
\usepackage{subcaption}
\usepackage{silence}
\newcommand{\xmark}{\ding{55}}
\newcommand{\cmark}{\ding{51}}
\setlist[itemize]{leftmargin=4.mm}

\usepackage[accsupp]{axessibility}  

\usepackage[pagebackref=true,breaklinks=true,letterpaper=true,colorlinks,bookmarks=false]{hyperref}

\iccvfinalcopy 

\begin{document}

\title{Visual Relationship Detection Using \\ Part-and-Sum Transformers with Composite Queries}

\author{Qi Dong \ \ Zhuowen Tu \ \ Haofu Liao  \ \ Yuting Zhang \ \ Vijay Mahadevan \ \ Stefano Soatto\\
Amazon Web Services \\
{\tt\small \{qdon,ztu,liahaofu,yutingzh,vmahad,soattos\}@amazon.com}
}

\maketitle
\ificcvfinal\thispagestyle{empty}\fi

\begin{abstract}

Computer vision applications such as visual relationship detection and human object interaction can be formulated as a composite (structured) set detection problem 
in which both the parts (subject, object, and predicate) and the sum (triplet as a whole) are to be detected in a hierarchical fashion. In this paper, we present a new approach, denoted Part-and-Sum detection Transformer (PST), to perform end-to-end visual composite set detection. Different from existing Transformers in which queries are at a single level, we simultaneously model the joint part and sum hypotheses/interactions with composite queries and attention modules. We explicitly incorporate sum queries to enable better modeling of the part-and-sum relations that are absent in the standard Transformers. Our approach also uses novel tensor-based part queries and vector-based sum queries, and models their joint interaction. We report experiments on two vision tasks, visual relationship detection and human object interaction and demonstrate that PST achieves state of the art results among single-stage models, while nearly matching the results of custom designed two-stage models.

\end{abstract}

\section{Introduction}

In this paper, we study problems such as {\em visual relationship detection} (VRD) \cite{lu2016visual,krishna2017visual} and {\em human object interaction} (HOI) \cite{gao2018ican,shen2018scaling,chao2018learning}
where a composite set of a two-level (part-and-sum) hierarchy is to be detected and localized in an image. In both VRD and HOI, the output consists of a set of entities. Each entity, referred to as a ``sum", represents a triplet structure composed of parts: the parts are ({\em subject, object, predicate}) in VRD and ({\em human, interaction, object}) in HOI. The sum-and-parts structure naturally forms a two-level hierarchical output - with the sum at the root level and the parts at the leaf level. In the general setting for composite set detection, the hierarchy consists of two levels, but the number of parts can be arbitrary.

\begin{figure}[!htp]
    \centering
    \scalebox{1.0}{
    \begin{tabular}{c}
 \hspace{-3mm}   \includegraphics[width=1.01\linewidth]{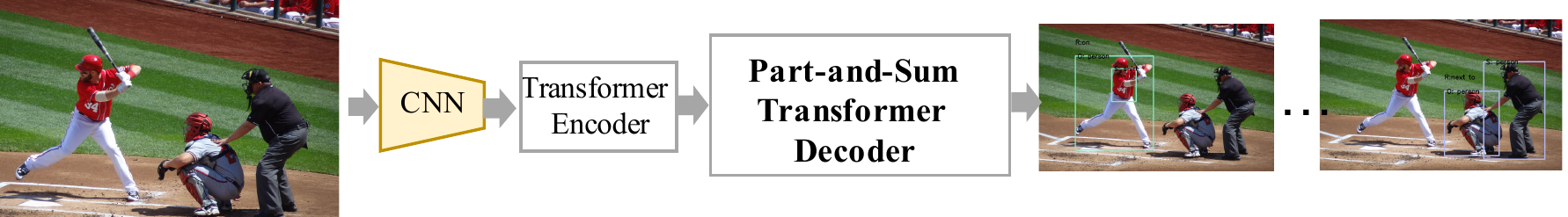} \\
    {\small (a) Pipeline overview.}\\
  \hspace{-3mm}  \includegraphics[width=1.01\linewidth]{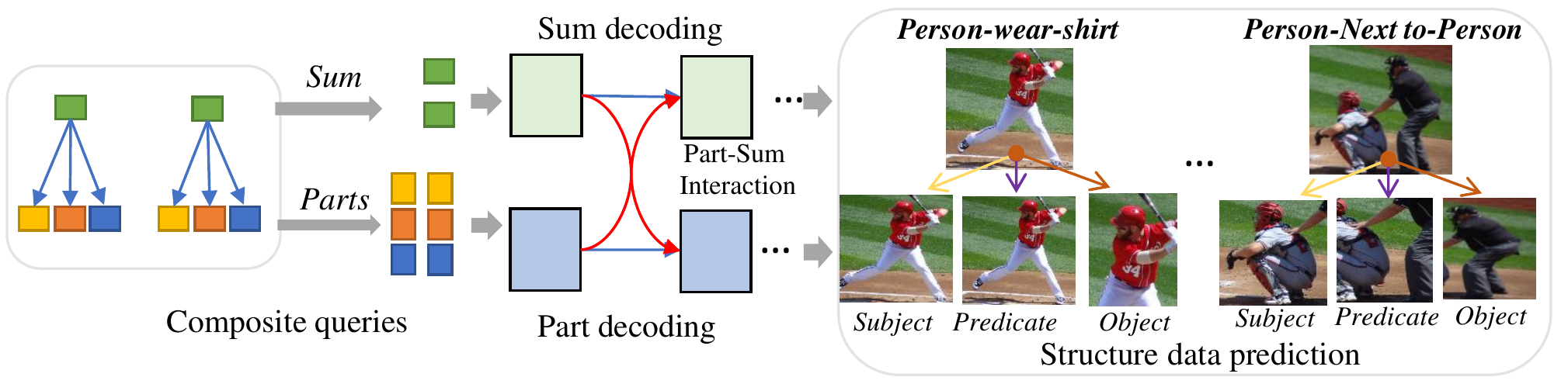} \\
    {\small (b) Part-and-Sum Transformer Decoder.}
    \end{tabular}
    }
    \vspace{-3mm}
    \caption{\small Overview of visual composite set detection by Part-and-Sum Transformer (PST).}
\label{fig:overview1}
\vspace{-5mm}
\end{figure}

Many existing approaches in VRD \cite{lu2016visual,hu2019neural,mi2020hierarchical,zhang2019graphical,zhang2019graphical} and HOI \cite{8354279,chao2018learning,li2020pastanet,hou2020visual,gao2020drg} are based on two stage processes in which some parts (e.g. the subject and the object in VRD) are detected first, followed by detection of the association (sum) and the additional part (the predicate). Single stage approaches \cite{yu2017visual,yin2018zoom,hu2013recognising,kim2020uniondet} with end-to-end learning for VRD and HOI also exist. In practice, two-stage approaches produce better performance while single-stage methods are easier to train and use.

The task for an object detector is to detect and localize all valid objects in an input image, making the output a set. Though object detectors such as FasterRCNN \cite{ren2015faster} are considered end-to-end trainable, they perform instance level predictions, and require post-processing using non-maximum suppression to recover the entire set of objects in an image. Recent developments in Transformers \cite{vaswani2017attention} and their extensions to object detection \cite{carion2020end} enable set-level end-to-end learning by eliminating anchor proposals and non-maximum suppression.

\begin{figure*}[!tp]
\centering
\scalebox{1.0}{
\begin{tabular}{ccc}
\includegraphics[width=0.8\linewidth]{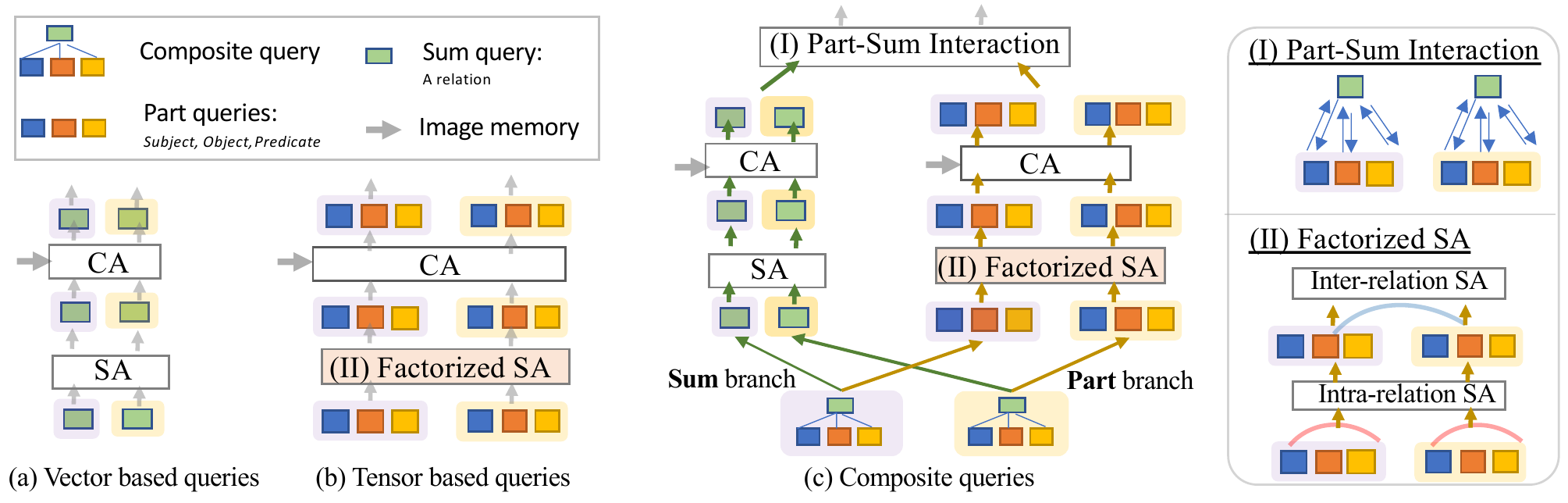}
\end{tabular}
}
\vspace{-2mm}
\caption{\small Part-and-Sum Transformer Decoder with Composite queries. We compare various Transformer with various query designs. Note that CA: Cross-attention layer; SA: Self-attention layer.}
\label{fig:partsum}
\vspace{-4mm}
\end{figure*}

In this paper, we formulate the visual relationship detection (VRD) \cite{lu2016visual,krishna2017visual} and human object interaction (HOI) \cite{gao2018ican,shen2018scaling,chao2018learning} as composite set (two-level hierarchy) detection problems and propose a new approach, part-and-sum Transformers (PST) to solve them. 
PST is different from existing detection transformers where each object is represented by a vector-based query in either a one-level set \cite{carion2020end}. We show the importance of establishing an explicit ``sum" representation for the triplet as a whole to be simultaneously modeled/engaged with the part queries (e.g., subject, object, and predicate). 
Both the global and the part features have also been modeled in the discriminatively trained part-based model (DPM) algorithm \cite{felzenszwalb2009object} though the global and part interactions there are limited to relative spatial locations. 
To summarize, we develop a new approach, part-and-sum Transformers (PST), to solve the composite set detection problem by creating composite queries and composite attention mechanism to account for both the sum (vector-based query) and parts (tensor-based queries) representations. Using systematic experiments, we study the roles of the sum and part queries and intra- and inter-token attention in Transformers. The effectiveness of the proposed PST algorithm is demonstrated in the VRD and HOI tasks.

\section{Related Work}

The standard object detection task \cite{lin2014microsoft} is a set prediction problem in which each element refers to a bounding box. If the prediction output is a hierarchy of multiple layers, e.g. $person\rightarrow face \rightarrow nose$, standard sliding window based approaches \cite{ren2015faster} that rely on features extracted from the entire window no longer suffice. 
Algorithms performing inference hierarchically exist \cite{zhu2007stochastic,serre2007feedforward,si2013learning} but they are not trained end-to-end. Here we study problems that require predictions of a two-level structured set including {\em visual relationship detection} (VRD) \cite{lu2016visual,krishna2017visual} and {\em human object interaction} (HOI) \cite{gao2018ican,shen2018scaling,chao2018learning}.
We aim to develop a general-purpose algorithm that is trained end-to-end for composite set detection with a new Transformer design, which is different from the previous two-stage \cite{lu2016visual,hu2019neural,mi2020hierarchical,zhang2019graphical,zhang2019graphical,8354279,chao2018learning,li2020pastanet,hou2020visual,gao2020drg} and single-stage approaches \cite{yu2017visual,yin2018zoom,hu2013recognising,kim2020uniondet}.
The concept of sum-and-max \cite{serre2007feedforward} is related to our part-and-sum approach but the two approaches have large differences in many aspects.
In terms of structural modeling, the problem of structured prediction (or semantic labeling) has been long studied in machine learning \cite{lafferty2001conditional,tsochantaridis2005large} and computer vision \cite{shotton2006textonboost,tu2008auto}.

Transformers~\cite{vaswani2017attention} have recently been applied to many computer-vision tasks~\cite{ vit,lxmert,touvron2020training,xu2021line,khan2021transformers}. Prominently, the object detector based on transformer (DETR)~\cite{carion2020end} has shown comparable results to well-established non-fully differentiable models based on CNNs and NMS, such as Faster RCNN~\cite{ren2015faster}. Deformable DETR~\cite{zhu2020deformable} has matched the performance of the previous state-of-the-art while preserving the end-to-end differentiability of DETR.
A recent attempt \cite{zou2021_hoitrans} also applies DETR to the HOI task. Our proposed Part-and-Sum Transformers (PST) is different in the algorithm design with the development of composite queries and composite attention to simultaneously model the global and local information, as well as their interactions.

\section{Part-and-Sum Transformers for Visual Relationship Detection}

In this section, we describe the PST formulation for visual composite set detection. We use VRD as the example, and the formulation can be straightforwardly extended to HOI by assuming that the subject is always {\em human} and the predicate as the {\em interaction} with the object. 

Given an image $I$, the goal of VRD is to detect a set of visual relations $S=\{R_i\}^{N}_{i=1}$. Each visual relation $R_i$, a {\em sum}, has three {\em parts}: subject, object and predicate, i.e., $R_i=\{s_i, p_i, o_i\}$. For each $R_i$, the subject and object have class labels $s_i$ and $o_i$ and bounding boxes $s^{b}_i$ and $o^{b}_i$. The predicate has a class label $p_i$. VRD is therefore a composite set detection task, where each instance in the set is a composite entity consisting of three parts.  

\subsection{Overview}

The overview of the proposed PST is shown in Figure~\ref{fig:overview1} (a). Given an input image, we first obtain the image feature maps from a CNN backbone. The image features with learnable position embeddings are further encoded/tokenized by a standard~\cite{carion2020end} or deformable~\cite{zhu2020deformable} transformer encoder. Those tokenized image features and a set of learnable queries are put into a transformer decoder to infer the classes and positions of every composite data. 
Unlike standard object detection, composite set detection not only detects all object entities but also entity-wise structure/relationships. For accurate modeling composite data, we propose composite query based Part-and-Sum transformer decoder, to learn each composite/structure data on both entity and relationship levels, as shown in Figure~\ref{fig:overview1} (b). In the following sections, we detail the PST model and the corresponding processes of training and inference.

\subsection{Part-and-Sum Transformer (PST)}
To construct the PST model, we first describe the vector based query, tensor based query, and composite queries that are used for composite set prediction. We then formulate the composite transformer decoder layer based on the composite queries.  
\subsubsection{Query Design for Composite Data}
 
 {\noindent\bf{Vector based query}} A standard decoder used in DETR takes a set of vector based queries as input, as shown in Figure~\ref{fig:partsum} (a). When applying this formulation to the relationship detection task, one can use feedforward networks (FFNs) to directly predict a subject-predicate-object triplet from the output of each query. This straightforward extension of DETR, while serving as a reasonable baseline, is sub-optimal as each query mixes the parts and their interaction altogether inside a vector. This makes the parts and their interaction implicitly modeled, limiting the expressiveness and representation capability of the visual relationship model.

{\noindent\bf{Tensor based query}} To explicitly model the parts and their relationships (e.g. Subject, Predicate, and Object), we propose a tensor-based query representation using disjoint sub-vectors as sub-queries.
Specifically for VRD, in a tensor based query representation, three sub-queries represent Subject, Predicate, and Object. All queries together form a $M \times P \times D$ matrix, where $M$ is the number of queries, $P$ is the number of entities in a relationship ($P=3$), and $D$ is the feature dimension of sub-queries.   This formulation enables part-wise decoding in the transformer decoder, as shown in Figure~\ref{fig:partsum} (b). Technically, the vector based query represents each relationship as a whole/Sum, whereas tensor based query models the parts disjointly. The difference in the query design leads to a difference in the learned contexts: self-attention layers among vector based queries mine inter-relation context, while self-attention layers among part queries mine the inter-part context.

 {\noindent\bf{Composite query}} 
 On the one hand, vector based query is able to capture the relationship as a sum/whole, but there exists an intrinsic ambiguity in the parts.
 On the other hand, tensor based query models each part explicitly, but it lacks the knowledge of the relationships as a sum, which is important for the subject-object association.
 Based on the observation above, we propose a \textit{composite} query representation. Formally, each composite query $Q_i$ is composed of part queries $Q_i^{P}$ (a tensor query) as well as a sum/whole query $q_i^{G}$ (a vector based query). In VRD, each composite query $Q_i$ is composed of three sub-queries to represent Subject, Predicate, and Object; and one sum query to represent the relationship. $Q_i=\{Q_i^{P}, q_i^{G}\}$, and $Q_i^{P}=\{q_i^s, q_i^p, q_i^o\}$, where $q_i^s$, $q_i^p$ and $q_i^o$ denote subject, predicate and object sub-queries. Assuming $M$ composite queries in the decoder, the overall query is a $M \times D\times4$ tensor, where $D$ is the dimension of sub-queries.

\subsubsection{Part-and-Sum Transformer Decoder}
\label{sec:PSTdecoder}
As the composite query includes both part and sum queries, we separately decode the part queries $Q^{P}$ and the sum query $q^{G}$. To enable mutual benefits of part and sum learning, we also set up part-sum interaction. Additionally, we propose a factorized self-attention layer for further enhancing part level learning. The architecture of Part-and-Sum Transformer Decoder is illustrated in Figure \ref{fig:partsum} (c).

{\noindent\bf{Part-and-Sum separate decoding}}. The PST decoder has a two-stream architecture, for part and sum queries decoding, respectively. Each decoding stream contains a self-attention layers (SA), cross-attention layers (CA), and feed-forward neural networks (FFN). Let $f$ and $\varphi$ denote respectively SA and CA layers for part queries. Decoding the part queries is written as:
\begin{align}
\begin{split}
    f_{\text{Part}}(\boldsymbol{Q}^{P}) & =\operatorname{SA}(\boldsymbol{Q}^{P}_{1},...,\boldsymbol{Q}^{P}_{M})\\
    & =\operatorname{SA}(q^{s}_{1},q^{p}_{1},q^{o}_{1},...,q^{s}_{M},q^{p}_{M},q^{o}_{M}) \\
     \varphi_{\text{Part}}(\boldsymbol{Q}^{P},\boldsymbol{I}) & =\operatorname{CA}([\boldsymbol{Q}^{P}_{1},...,\boldsymbol{Q}^{P}_{M}],\boldsymbol{I}),
    \end{split}
    \label{eq:part1}
\end{align}
where $\boldsymbol{I}$ denotes the tokenized image features from the Transformer Encoder. 
Similarly, decoding the sum queries can be written as:
\begin{align}
\begin{split}
   f_{\text{Sum}}(\boldsymbol{Q}^{G}) & =\operatorname{SA}(q^{G}_{1},...,q^{G}_{M})\\
    \varphi_{\text{Sum}}(\boldsymbol{Q}^{G},\boldsymbol{I}) & =\operatorname{CA}([q^{G}_{1},...,q^{G}_{M}],\boldsymbol{I})
    \end{split}
    \label{eq:global1}
\end{align}
Each entity has both part and global embeddings via two separate sequential modules $\text{FFN}(\varphi(f(\boldsymbol{Q}),\boldsymbol{I}))$. The self-attention exploits the context among all queries. Part-and-Sum separated decoding effectively models two different types of contexts: the self-attention in part queries explores the inter-component context, for example, when one part query predicts ``person", it reinforces related predicates such as ``eat" and ``hold"; while self-attention for global queries exploits the inter-relationship context, for example, a sum query that predicts ``Person read book", is a clue to infer ``person sit" relationship. These contexts provide the interactions needed for the accurate inference of the structured output.

{\noindent\bf{Factorized self-attention layer}}. To make the interactions within a group-wise part query more structured, we design a factorized self-attention layer, as shown in Figure \ref{fig:partsum} (b). Instead of doing self-attention among all part queries as in Eq.~\ref{eq:part1}, a factorized self-attention layer first conducts intra-relation self-attention, and then conducts inter-relation self-attention. The intra-relation self attention layer leverages the contexts of the parts to benefit relationship prediction, for example, subject query and object query are ``person" and ``horse" helps predict predicate ``Ride". The inter-relation self-attention layer leverages the inter-relation context, to enhance the holistic relation prediction per image, which is particularly important for multiple interactions detection for the same subject entity. More details are in the supplementary materials.

{\noindent\bf{Part-Sum interaction}}. Part query decoding embeds more accurate component information, while global embedding contains more accurate component association. These two aspects are both important to the structured output detection, and are mutually beneficial to each other \cite{felzenszwalb2010cascade}. Thus, we design interaction between the two decoding streams, enabling part-sum conditions. Specifically, after FFNs in the decoder, for each part embedding $q^{k}_{i}, k\in\{s,o,p\}$, we combine it with the sum query embedding, while for each sum query $q^{G}_{i}$, we fuse all three part query embeddings. The part-sum interaction is formulated as:
\begin{equation}
 \begin{split}
   &q^{k}_{i}=\mathcal{N}(q^{k}_{i}+q^{G}_{i}), k\in\{s,o,p\} \\
   & q^{G}_{i}=\mathcal{N}(q^{G}_{i}+\sum_{k\in\{s,o,p\}}q^{k}_{i}) ,
    \label{eq:interaction}
    \end{split}
\end{equation}
where $\mathcal{N}$ is layer normalisation \cite{ba2016layer}.

\subsection{Model Training and Inference}
{\noindent\bf{Composite prediction}}. For each composite query $Q_{i}=\{q^{G}_{i}, q^{s}_{i},q^{o}_{i}, q^{p}_{i}\}$, we predict the classes of subject, object and predicate; and the bounding boxes for subject and object. Specifically, for each part query, we predict corresponding classes by using a one-layer linear layer, and predict boxes using a shallow MLP head. Besides, we can also construct the global representation from group-wise part queries, by concatenating all part queries, and denote this as $q^{\text{spo}}_{i} = [q^{s}_{i},q^{o}_{i}, q^{p}_{i}]$. The part query predication is:     
\begin{equation}
    \begin{split}
    &\hat{b}_i^k = f^{k}_{\text{box}}(q_i^k), k\in\{s,o\}\\
    &\hat{p}_i^k = f^k_{\text{cls}}(q_i^k), k\in\{s,o,p\} \\
    &\hat{p}_i^{\text{spo}} = f^{\text{spo}}_{\text{cls}}(q_i^{\text{spo}})
    \label{eq:relation_prediction_local}
    \end{split}
\end{equation}
where $f^{*}_{\text{cls}}$ are the FFNs for subject, object and predicate classification; and $f^{*}_{\text{box}}$ are the FFNs for predicting the boxes of subject and object; $f^{\text{spo}}_{\text{cls}}$ is an FFN for predicting relation triplet. 
 
For Sum query prediction, we predict classes and boxes of all parts from a Sum query $q_i^{G}$, that is:
\begin{equation}
    \begin{split}
    \hat{b}_i^G = g^{k}_{\text{box}}(q_i^\text{G}), k\in\{s,o\}\\
    \hat{p}_i^G = g^{k}_{\text{cls}}(q_i^\text{G}), k\in\{s,o,p\}
    \label{eq:relation_prediction_global}
    \end{split}
\end{equation}
where $g^{*}_{\text{cls}}$ are FFNs for subject, object and predicate classification; and $g^{*}_{\text{box}}$ are the FFNs for predicting the boxes of subject and object in the global level. 
Note that the last layer in $f_{\text{cls}}$ is a Softmax layer, while the last layer in $f_{\text{box}}$ is a Sigmoid layer. 

{\noindent\bf{Composite bipartite matching}}.
We conduct composite group-wise bipartite matching, i.e. considering all components belonging to a relation jointly in computing set-to-set similarity. Specifically, for a relation, there are three Part queries (subject, object and predicate), and a triplet embedding.
The bipartite matching algorithm finds a permutation $\sigma$ of the $M$ predictions $\{\hat{y}_i\}_{i=1}^M$ so that the total matching cost is minimized
\begin{equation}
\vspace{-1mm}
    \hat{\sigma} = \underset{\sigma \in \mathcal{P}}{\arg \min} \sum_{i}^{M}{\mathcal{C}_{\text{match}}}(y_i, \hat{y}_{\sigma(i)}),
\vspace{-1mm}
\end{equation}
where $\mathcal{P}$ is a set of all possible permutation of $M$ elements, and $\sigma(i)$ is the $i$th element of the permutation $\sigma$.

We define the following matching cost between the $i$th ground truth $y_i$ and the corresponding $i$th prediction determined by permutation $\sigma$:
\begin{equation}
    \begin{split}
    \mathcal{C}_{\text{match}}&(y_i, \hat{y}_{\sigma(i)}) = \mathcal{C}^{\text{Part}}_{\text{match}}(y_i, \hat{y}_{\sigma(i)})+\mathcal{C}^{\text{Sum}}_{\text{match}}(y_i, \hat{y}_{\sigma(i)}) \\
    & =\sum_{t \in \{s, p, o, \text{spo},\text{G}_s, \text{G}_p, \text{G}_o\}} -\mathds{1}_{\{c_i^{t} \ne \varnothing\}}\hat{p}^t_{\sigma(i)}(c_i^{t})\\
    & + \sum_{t \in \{s, p, o, \text{G}_s, \text{G}_o\}}\mathds{1}_{\{c_i^{t} \ne \varnothing\}} \mathcal{L}_{\text{box}}(b_i^t, \hat{b}_{\sigma(i)}^t).
    \end{split}
    \label{eq:matching_cost}
\end{equation}
\vspace{-0.1cm}
where $\hat{p}^t_{\sigma(i)}(c_i^{t})$ is the probability of classifying $t$ as $c_i^{t}$ computed by Eq. \ref{eq:relation_prediction_local} and \ref{eq:relation_prediction_global}, and $\hat{b}_{\sigma(i)}^t$ is a predicted bounding box ($\text{G}_s$ and $\text{G}_o$ denote the subject and object embeddings from a Sum query branch). $\mathcal{L}_{\text{box}}$ includes GIoU and L1 losses, same as \cite{carion2020end}. Here we use a union box of the subject and object in a relation to represent the target box of a corresponding predicate.

{\noindent\bf{Training loss}}. Given two-level Part and Sum outputs, we compute classification losses and box regression losses on both levels. Once we obtained the best permutation $\sigma$ that minimizes the overall matching cost between $y$ and $\hat{y}$, we can compute the total loss as
\begin{equation}
\small
    \begin{split}
    &\mathcal{L}(y, \hat{y}) = \sum_{i=1}^{M}\big(\mathcal{L}_{\text{Part}}(y, \hat{y})+\mathcal{L}_{\text{Sum}}(y, \hat{y})\big)\\
    &\mathcal{L}_{\text{Part}}(y, \hat{y}) = 
    \sum_{t \in \{s, p, o,\small\text{spo}\}} -\log{\hat{p}^t_{\sigma(i)}(c_i^{t})}
    +   \mathcal{L}_{\text{box}}(b_i^t, \hat{b}_{\sigma(i)}^t)) \\
    &\mathcal{L}_{\text{Sum}}(y, \hat{y}) = \sum_{t \in \{\text{G}_{s}, \text{G}_{p}, \text{G}_{o}\}} -\log{\hat{p}^t_{\sigma(i)}(c_i^{t})})
    +\mathcal{L}_{\text{box}}(b_i^t, \hat{b}_{\sigma(i)}^t) 
    \end{split}
    \label{eq:relation_loss}
\end{equation}
Note that Eq. \ref{eq:relation_loss} is very similar to Eq. \ref{eq:matching_cost}, except that a negative log-likelihood loss is used to train classifiers, for more effective learning.

\begin{table*}[!ht] 
\centering
\setlength{\tabcolsep}{0.43cm}
\caption{\small Phrase and relationship detection result comparison (\%) under various Transformer architectures on the VRD dataset. For a fair comparison, all transformer decoders are based on the same CNN backbone and Transformer encoder, using the same number of queries.}
\vspace{-2mm}
\label{tab:architecture}
\scalebox{0.6}{
\begin{tabular}{cccc|ccc|cccccccc}
\hline
&\multicolumn{3}{c|}{Query Type}&\multicolumn{3}{c|}{Transformer Decoder Design} & \multicolumn{4}{c}{Phrase Detection}& \multicolumn{4}{c}{Relationship Detection}\\ 
No.&Vanilla &Tensor&Composite&Vanilla&Part & Part-and-Sum & \multicolumn{2}{c}{$k=1$}& \multicolumn{2}{c}{$k=70$}& \multicolumn{2}{c}{$k=1$}& \multicolumn{2}{c}{$k=70$}\\
&&&&&&& R@50 & R@100 & R@50 & R@100 & R@50 & R@100 & R@50 & R@100  \\\hline  
(a)&\cmark&&&\cmark&& &26.17&29.43&27.66&32.71&17.88&19.41&19.97&23.08\\ 
(b)&&\cmark&&\cmark&& &26.69&31.46&28.67&34.35&19.36&22.63&21.89&25.89\\
(c)&&\cmark&&&\cmark& &30.40& \bf 34.86& 32.29& 37.68& 23.28&26.30& 25.46& 29.65 \\
(d)&&&\cmark&\cmark& & & 25.70&29.66&28.01&34.11&17.75&20.20&20.17&24.53\\
(e)&&&\cmark&& & \cmark& \textbf{30.63} & 33.82 & \textbf{32.55}&\textbf{40.63} &\textbf{23.57}& \textbf{27.63}& \textbf{26.48}& \textbf{31.83}\\
\hline 
\end{tabular}
}
\end{table*}

\section{Experiments}
We evaluate our method on two composite set detection applications: Visual Relationship Detection ({\bf VRD}), and Human Object Interaction detection ({\bf HOI}).

{\noindent\bf Datasets.} (1) For the VRD task, we evaluate the proposed PST on VRD dataset \cite{lu2016visual}, containing $5{,}000$ images, and $100$ entity categories and $70$ predicate categories. Relationships are labeled as a set of $<${\em subject, predicate, object}$>$ triplets, and all subject and object entities in relationships are annotated with an entity category and a bounding box. We follow the data split of \cite{lu2016visual} and use $3{,}700/300/1{,}000$ images for training/validation/test. There are $37{,}993$ visual relationship instances that belong to $6{,}672$ triplet types, and $1{,}169$ relation types that only appear in the test set, which are used for zero-shot relationship detection.
(2) For HOI task, we conduct an evaluation on HICO-DET dataset \cite{chao2018learning}, including 38,118 training images and 9,658 testing images. In this dataset, there are the same 80 object categories as MS-COCO \cite{lin2014microsoft} and 117 verb categories, and objects and verbs construct 600 classes of HOI triplets. One person is able to interact with multiple objects in various ways at the same time in this dataset. 

{\noindent\bf Task Settings.} For the VRD task, we test PST on Phrase Detection and Relationship Detection \cite{lu2016visual, zhang2019graphical}. In Phrase Detection, the model detects one bounding box for each relationship, and recognizes the categories of subject, object and predicate in the relationship. In Relationship Detection, the model detects two individual bounding boxes for both subject and object entities in the relationship, and classifies the subject, predicate and object in the relationship. In both tasks, we consider two settings: single and multiple predicates between a pair of subject and object entities, with $k$ representing the number of predicates between a pair. 

{\noindent\bf Performance Metrics.} (1) In VRD, we use relationship detection recall@K as the evaluation metric, as the true relationships annotations are incomplete. Following the evaluation in \cite{lu2016visual}, for each detected relationship, we compute joint probability of subject, predicate and object category prediction as the score for that relationship,
and then rank all detected relationships to compute the recall metric. For a relationship to be detected correctly, all three elements are required to be correctly classified and the IoU between the predicted bounding boxes and groundtruth bounding boxes are greater than 0.5. (2) In HOI, we use mean average precision (mAP) \cite{gao2018ican} as the evaluation metric. An HOI detection is considered correct only when the action and the object class are both correctly recognized and the corresponding human and object bounding boxes detection have higher than 0.5 IoU with the groundtruth boxes.

{\noindent\bf Implementation Details.} On both VRD and HOI task, PST shares configurations. PST uses the standard ResNet-50 network as the backbone, followed by a Transformer encoder with six encoder layers, the same as Deformable DETR \cite{zhu2020deformable}. The proposed PST decoder contains six layers of the proposed two-stream Part-and-Sum decoder layers. All feed-forward networks are two-linear-layer shallow networks. We set up auxiliary losses after each decoder layer, and use 400 composite queries, with three part queries representing the subject, object and predicate in a visual relation or human, object and interaction in a HOI, respectively. Note that, in our experiment, we use the vanilla multi-head self-attention module \cite{vaswani2017attention} as a self-attention layer, and use a deformable multi-head cross-attention module as a cross-attention layer. More details are in the supplementary.


\subsection{Part-and-Sum Transformer Analysis}

{\noindent{\bf Part-and-Sum Transformer Decoder}} We first compare and analyze different Transformer designs, with different query types on VRD. The various Transformer designs are compared in Figure \ref{fig:partsum}. Vector based query is the most straightforward way to detect structured outputs using a Transformer, by formulating an individual structure entity as a vector query, and feeding queries into a vanilla Transformer decoder \cite{zhu2020deformable} to learn embeddings for each relation. Then, three one-linear-layer heads are used to predict the classes of subjects, objects and predicates, and two three-layer MLP heads to regress the boxes. The result comparison are shown in Table \ref{tab:architecture}. 

We can see that (1) with vanilla transformer decoder, tensor based query outperforms vector based query (in (a) vs (b)), with the margin 1.01/1.64\% and 2.03/2.81\% at R@50/100 on Phrase and Relationship detection ($k=70$). It is because vector query models the structure entity as a whole, and embeds multiple parts in one query. This design increases the difficulty of Hungarian matching. (2) For Tensor based query, Part Transformer outperforms the Vanilla Transformer with a clear margin (in (b) vs (c)). This benefit mainly comes from the factorized design in the self-attention layer, and the relation-level constraint (in Eq. \ref{eq:relation_loss}). The former enhances the intra-relation context, reducing the ambiguity of entity recognition, for example, Subject ``Person" and Object ``horse" are important clues to infer Predicate ``Ride". The latter learns the relation as a whole to reduce the entity instance confusion \cite{zhang2019graphical}. (3) Despite the part-and-sum benefit inside the composite query, vanilla transformer with composite query degrades (in (d) vs (a)), compared with using the vanilla query. It shows that directly mixing Part and Sum queries cannot benefit structured output learning, because per Sum query contains multiple parts, and some relations may share the same entity instance, which could confuse the similarity computation in self-attention modules. To leverage two-level information and context effectively, PST decodes Part and Sum queries separately, with group-wise part-sum interaction. By comparing (d) vs (e), this design outperforms vanilla transformer, with the margin 4.54/6.52\% and 6.31/7.30\% at R@50/100 on Phrase and Relationship detection ($k=70$).     

{\noindent{\bf Factorized self-attention layer}}
We check the effectiveness of the factorization self-attention layer in the Part query decoding stream. We compare the performance of PST with factorization self-attention layer vs PST with vanilla self-attention layer on VRD, and the results are shown in Table \ref{tab:ab1}. It shows that factorized self-attention design leads to 1.18/2.66\% improvement at R@50/100 on Relationship detection ($k=70$). 

{\noindent{\bf Part-Sum interaction}} We compare two Part-Sum interaction schemes: Vanilla Self-attention vs Summation operation. The results are shown in Table \ref{tab:ab1}. From it, we can see that Part-Sum bidirectional summation works better than self-attention interaction, mainly due to the determined grouping configuration between part and sum queries in PST as shown in Figure \ref{fig:partsum}(c) \footnote{In PST, the component order of a composite query is fixed. On VRD, for instance, the first part query is for Subject, the second for Object, and the third for Predicate. The grouping between part queries and sum query is also fixed by design, i.e. the first sum query and the first group of part queries represent the same relation instance.}. 

\begin{table}[!ht] 
\centering
\setlength{\tabcolsep}{0.4cm}
\caption{\small Ablation study on Part-Sum Transformer designs. We report Relationship detection result comparison (\%) on VRD.}
\vspace{-3mm}
\label{tab:ab1}
\scalebox{0.6}{
\begin{tabular}{cc|cccc}
\hline
&& \multicolumn{4}{|c}{Relationship Detection}\\ 
{Module}& &\multicolumn{2}{c}{$k=1$}& \multicolumn{2}{c}{$k=70$}\\
&& R@50 & R@100 & R@50 & R@100  \\\hline
\multirow{2}{*}{Factorization SA}& \xmark& 22.14&26.48&25.30&29.17\\
&\cmark&\bf{23.57}& \bf{27.63}& \bf{26.48}& \bf{31.83}\\ \hline
\multirow{2}{*}{Part-Sum Interaction}&
Self-Attention&22.04&25.42&23.89&28.87\\
&Part$\leftrightarrow$Sum&\bf{23.57}& \bf{27.63}& \bf{26.48}& \bf{31.83}\\\hline
\end{tabular}
} 
\vspace{-3mm}
\end{table}

{\noindent{\bf Composite prediction}} Given the two-stream design of Part-and-Sum decoding, we obtain the predictions from both part and sum levels. Thus, we study the various inference schemes: predicting structure data from part query branch, or from sum query branch, or combining two branches. To combine predictions from part and sum queries, for classification probability, we average the prediction probability of group-wise part and sum queries; for box prediction, we just average predicted positions of left-top and right-bottom points. The results comparison are shown in Table \ref{tab:inference}, and it shows that Part only inference slightly outperforms Sum only inference, and combining the predictions from two levels is able to bring a minor improvement for relationship detection \footnote{We report the results by part query only based inference in both VRD and HOI experiments for clarity.}.

\begin{table}[!ht] 
\centering
\vspace{-2mm}
\setlength{\tabcolsep}{0.4cm}
\caption{\small Result comparison of various inference schemes.}
\vspace{-2mm}
\label{tab:inference}
\scalebox{0.75}{
\begin{tabular}{ccccc}
\hline
& \multicolumn{4}{c}{Relationship Detection}\\ 
Inference& \multicolumn{2}{c}{$k=1$}& \multicolumn{2}{c}{$k=70$}\\
& R@50 & R@100 & R@50 & R@100  \\\hline
Part only&23.57& \bf{27.63}& 26.48& 31.83\\ 
Sum Only&22.06&25.43&25.76&30.45\\
Part-Sum&\bf{24.34}&27.01&\bf{27.03}&\bf{31.90}  \\\hline
\end{tabular}
}  
\vspace{-5mm}
\end{table}

\subsection{Visual Relationship Detection}
We compare PST with existing visual relationship detection solutions on the VRD datasets \cite{lu2016visual}.

{\noindent\bf Competitors.} 
Existing visual relationship detection solutions can be classified into two categories: 
{\em(I) Stage-wise methods}: These approaches first detect objects using pre-trained detector, and then use the outputs of the object detector
as {\em fixed} inputs of a relationship detection module. Specifically, we compare with: (1) VRD-Full \cite{lu2016visual} which combines the visual appearance and the language features of candidates boxes to learn relationships. 
(2) NMP \cite{hu2019neural} which builds a relationship graph and optimizes it by node-to-edge and edge-to-node message passing mechanisms.
(3) CDDN \cite{cui2018context} which proposes a context guided visual-semantic feature fusion scheme for predicate detection.
(4) LSVR \cite{zhang2019large} which learns a better representation by aligning the features on both entity and relationship levels.
(5) RelDN \cite{zhang2019graphical} which uses contrastive loss functions to learn fine-grained visual features. 
(6) BCVRD \cite{inayoshi2020bounding} which proposes a new box-wise fusion method to better combine visual, semantic and spatial features. 
(7) HGAT \cite{mi2020hierarchical} which proposes to use object-level and triplet-level reasoning to improve relationship detection.
{\noindent\em (II) End-to-End methods}: These approaches detect objects and relationships jointly. Specifically, we compare with:
(1) CAI \cite{zhuang2017towards} - leverages the subject-object context to detect relationships.
(2) KL distillation\cite{yu2017visual} - uses a linguistic model to regularize the visual model learning.
(3) DR-Net \cite{dai2017detecting} - designs a fully connected network to mine object-pair relationships.
(4) Zoom-Net \cite{yin2018zoom} - leverages multi-scaled relation contexts.
(5) VTransE \cite{zhang2017visual} - learns to map the visual features to the relationship space.  

\begin{table}[!tp] 
\centering
\setlength{\tabcolsep}{0.11cm}
\caption{\small Phrase and relationship detection result comparison (\%) on VRD dataset. - denotes that the results are not reported in the original paper. $k$ is the number of predicates associated with each subject-object pair. Note on VRD dataset, the maximum number of predicates is $k=70$. The {\em first} block is for the stage-wise detection methods, and the {\em second} block is for end-to-end detection methods. Our method belongs to the latter. $^\dagger$: the reported results of BC-VRD are based on Faster R-CNN for a fair comparison.}
\label{tab:vrd}
\scalebox{0.65}{
\begin{tabular}{c|cc cc|cccc}
\hline
Method & \multicolumn{4}{c|}{Phrase Detection}& \multicolumn{4}{c}{Relationship Detection}\\ 
& \multicolumn{2}{c}{$k=1$}& \multicolumn{2}{c|}{$k=70$}& \multicolumn{2}{c}{$k=1$}& \multicolumn{2}{c}{$k=70$}\\
& R@50 & R@100 & R@50 & R@100 & R@50 & R@100 & R@50 & R@100  \\\hline  
    VRD-Full \cite{lu2016visual}&16.17&17.03&20.04&24.90&13.86&14.70&17.35&21.51\\
    LSVR\cite{zhang2019large}&18.32&19.78&21.39&25.65&16.08&17.07&18.89&22.35\\
 BC-VRD \cite{inayoshi2020bounding}$^\dagger$&19.72&20.95&24.47&28.38  &15.87 &16.63&19.91 &22.86\\
    MLA-VRD \cite{zheng2019visual}&23.36&28.12&-&-&20.54&24.91&-&-\\
     NMP \cite{hu2019neural} &-&-&-&-&20.19&23.98&21.50&27.50\\
    HGAT \cite{mi2020hierarchical} &-&-&-&-&22.52&24.63&22.90&27.73\\ 
    RelDN-IMG \cite{zhang2019graphical}&26.37& 31.42&28.24&35.44&19.82&22.96&21.52&26.38\\
    MF-URLN \cite{zhan2019exploring} &{\bf31.50}&36.10&-&-&23.90 &26.80&-&-\\
    {\bf RelDN} \cite{zhang2019graphical}& 31.34 & {\bf 36.42}&{\bf 34.45}&{\bf 42.12}&{\bf 25.29}&{\bf 28.62}&{\bf 28.15}&{\bf 33.91}\\\hline
     DR-Net \cite{dai2017detecting} &-&-&19.93&23.45 &-&-&17.73&20.88\\
    VTransE \cite{zhang2017visual} &19.42&22.42&-&-&14.07&15.20&-&-\\
    CAI \cite{zhuang2017towards}&17.60&19.24&-&-&15.63&17.39&-&-\\
        ViP \cite{li2017vip}&22.80&27.90&-&-&17.30&20.00&-&-\\
    KL distilation\cite{yu2017visual}&23.14&24.03&26.32&29.43&19.17&21.34&22.68&31.89\\
    Zoom-Net \cite{yin2018zoom}&24.82&28.09&29.05&37.34&18.92&21.41&21.37&27.30\\
    {\bf PST} (ours) &{\bf30.63}&{\bf 33.82}&{\bf 32.55}&{\bf 40.63}&{\bf 23.57}&{\bf 27.63}&{\bf 26.48}&{\bf 31.83}\\
    \hline 
\end{tabular}
}
\vspace{-4mm}
\end{table}

{\noindent\bf Results.} The visual relation detection comparisons on VRD dataset are shown in Table \ref{tab:vrd}. For clarity, the stage-wise methods are grouped in the {\em first} block, and end-to-end methods are in the {\em second} block. The proposed PST belongs to the {\em second} block, and particularly it is the first holistic end-to-end VRD solution (directly outputs all predicted relationships without any post-processing). It is evident that PST outperforms the existing end-to-end methods on both Phrase and Relationship Detection tasks, e.g. surpassing the second best end-to-end method Zoom-Net \cite{zhan2019exploring} with a margin of 5.81$\%$/5.73$\%$ in Phrase Detection, and 4.65\%/6.22\% in Relationship detection at R@50/100 when $k=1$. 
It shows that PST is able to learn the relationships between all entities effectively.

\begin{table*} 
\centering
\setlength{\tabcolsep}{0.11cm}
\caption{\small Comparison with state-of-the-art HOI methods on HICO-DET dataset. For ``Detector", ``COCO" refers to an off-the-shelf object detector trained on COCO. ``HICO-DET" means the COCO pretrained object detector is further finetuned on HICO-DET. ``Pose" refers to using human pose as additional features. ``Language" refers to adopting the language priors.}

\label{tab:hico-det1}
\scalebox{0.8}{
\renewcommand{\arraystretch}{0.85}
\begin{tabular}{ccccc|ccc|ccc}
\hline
& & & & & \multicolumn{3}{c}{Default} & \multicolumn{3}{|c}{Known Object} \\
Method & Backbone & Detector & Pose & Language & Full$\uparrow$ & Rare$\uparrow$ & NonRare$\uparrow$ & Full$\uparrow$ & Rare$\uparrow$ & NonRare$\uparrow$ \\ \hline
\multicolumn{5}{l|}{\textit{Two-stage methods}} & & & & & & \\
Shen et al.\cite{8354279} & VGG19 & COCO & & & 6.46 & 4.24 & 7.12 & - & - & - \\
HO-RCNN\cite{chao2018learning} & CaffeNet & COCO & & & 7.81 & 5.37 & 8.54 & 10.41 & 8.94 & 10.85\\
InteractNet\cite{DBLP:journals/corr/GkioxariGDH17} & ResNet-50-FPN & COCO & & & 9.94 & 7.16 & 10.77 & - & - & - \\
GPNN\cite{DBLP:journals/corr/abs-1808-07962} & ResNet-101 & COCO & & & 13.11 & 9.34 & 14.23 & - & - & - \\
iCAN\cite{gao2018ican} & ResNet-50 & COCO & & & 14.84 & 10.45 & 16.15 & 16.26 & 11.33 & 17.73 \\
PMFNet-Base\cite{DBLP:journals/corr/abs-1909-08453} & ResNet-50-FPN & COCO & & & 14.92 & 11.42 & 15.96 & 18.83 & 15.30 & 19.89 \\
PMFNet\cite{DBLP:journals/corr/abs-1909-08453} & ResNet-50-FPN & COCO & \checkmark & & 17.46 & 15.65 & 18.00 & 20.34 & 17.47 & 21.20 \\
No-Frills\cite{DBLP:journals/corr/abs-1811-05967} & ResNet-152 & COCO & & \checkmark & 17.18 & 12.17 & 18.68 & - & - & - \\
TIN\cite{DBLP:journals/corr/abs-1811-08264} & ResNet-50 & COCO & \checkmark & & 17.22 & 13.51 & 18.32 & 19.38 & 15.38 & 20.57 \\
CHG\cite{wang2020contextual} & ResNet-50 & COCO & & & 17.57 & 16.85 & 17.78 & 21.00 & 20.74 & 21.08 \\
Peyre et al.\cite{peyre2019detecting} & ResNet-50-FPN & COCO & & \checkmark & 19.40 & 14.63 & 20.87 & - & - & - \\
IPNet \cite{wang2020learning} & Hourglass & COCO & & & 19.56 & 12.79 & 21.58 & 22.05 & 15.77 & 23.92 \\
VSGNet\cite{ulutan2020vsgnet} & ResNet-152 & COCO & & & 19.80 & 16.05 & 20.91 & - & - & - \\
FCMNet\cite{Liu20a} & ResNet-50 & COCO & \checkmark & \checkmark & 20.41 & 17.34 & 21.56 & 22.04 & 18.97 & 23.12 \\
ACP\cite{kim2020detecting} & ResNet-152 & COCO & \checkmark & \checkmark & 20.59 & 15.92 & 21.98 & - & - & - \\
Bansal et al.\cite{bansal2020detecting} & ResNet-50-FPN & HICO-DET & & \checkmark & 21.96 & 16.43 & 23.62 & - & - & - \\
PD-Net\cite{zhong2020polysemy} & ResNet-152 & COCO & & \checkmark & 20.81 & 15.90 & 22.28 & 24.78 & 18.88 & 26.54 \\
PastaNet\cite{li2020pastanet} & ResNet-50 & COCO & \checkmark & \checkmark & 22.65 & \textbf{21.17} & 23.09 & 24.53 & {23.00} & 24.99 \\
VCL\cite{hou2020visual} & ResNet-101 & HICO-DET & & & 23.63 & 17.21 & 25.55 & 25.98 & 19.12 & 28.03 \\
DRG\cite{gao2020drg} & ResNet-50-FPN & HICO-DET & & \checkmark & \textbf{24.53} & {19.47} & \textbf{26.04} & \textbf{27.98} & \textbf{23.11} & \textbf{29.43} \\
\hline
\multicolumn{5}{l|}{\textit{One-stage methods}} & & & & & & \\
UnionDet \cite{kim2020uniondet} & ResNet-50-FPN & HICO-DET & & & 17.58 & 11.52 & 19.33 & 19.76 & 14.68 & 21.27 \\
PPDM \cite{liao2020ppdm} & Hourglass & HICO-DET & & & 21.73 & 13.78 & 24.10 & 24.58 & 16.65 & 26.84 \\
HoiT \cite{zou2021_hoitrans} & ResNet-50 & - & & & {23.46} & \textbf{16.91} & 25.41 & 26.15 & \textbf{19.24} & 28.22 \\
\textbf{PST} (Ours) & ResNet-50 & - & & & \textbf{23.93} &{14.98} & \textbf{26.60} & \textbf{26.42} & {17.61} & \textbf{29.05} \\\hline 
\end{tabular}
}
\vspace{-1mm}
\end{table*}

From the comparisons with the stage-wise VRD methods, PST outperforms the second best method HGAT \cite{mi2020hierarchical} with a margin of 2.8\% at R@50 in the Relationship detection, but lags behind the best method RelDN \cite{zhang2019graphical} with a margin of 0.71\% and 1.32\% on Phrase and relationship detection at R@50 with $k=1$. We note that RelDN is a sophisticated two-stage method that:(1) leverages two CNNs for the entity and predicate visual feature learning; (2) tunes the thresholds of three margins in metric learning based losses functions; (3) combines multi-modality information (visual, semantic and spatial information) for relationship prediction. By contrast, PST predicts relations just based on visual features and detects relationships end-to-end and holistically without any post-processing. PST is simple, without any hand-designed components to represent the prior knowledge.

\subsection{Human Object Interaction Detection}

{\noindent\bf Competitors.} We compare our model with two types of state-of-the-art HOI methods: the two-stage methods
\cite{8354279,DBLP:journals/corr/GkioxariGDH17,DBLP:journals/corr/abs-1808-07962,gao2018ican,DBLP:journals/corr/abs-1909-08453,DBLP:journals/corr/abs-1811-05967,DBLP:journals/corr/abs-1811-08264,wang2020contextual,peyre2019detecting,ulutan2020vsgnet,Liu20a,kim2020detecting,bansal2020detecting,zhong2020polysemy,li2020pastanet,hou2020visual,gao2020drg,chao2018learning} and the single-stage methods \cite{kim2020uniondet,liao2020ppdm,zou2021_hoitrans}. The two-stage methods aim to detect individual objects in the first stage. Then, they associate the detected objects and infer the HOI predictions in the second stage. Two-stage methods rely on good object detections in the first stage and mostly focus on the second stage where language priors \cite{DBLP:journals/corr/abs-1811-05967,peyre2019detecting,Liu20a,kim2020detecting,bansal2020detecting,zhong2020polysemy,li2020pastanet,gao2020drg} and human pose features \cite{DBLP:journals/corr/abs-1909-08453,DBLP:journals/corr/abs-1811-08264,Liu20a,kim2020detecting,li2020pastanet} may be leveraged to facilitate the inference of HOI predictions from detected object. The single-stage methods aim to bypass the object detection step and directly output HOI predictions in one step. Previous single-stage methods \cite{kim2020uniondet,liao2020ppdm} are not end-to-end solutions. They employ multiple branches with each branch outputs complementary HOI-related predictions and rely on post-processing to decode the final HOI predictions. The most related to our approach is HoiT \cite{zou2021_hoitrans} which is an end-to-end single stage solution. HoiT employs DETR-like structure and predicts an HOI triplet from each vector query. 
\begin{table}[!th]
\centering
\caption{\small Comparison(\%) of a shared-stream vs independent-stream in the PST decoder. Shared-stream: Part and Sum queries share the same SA and CA layers; independent-stream: Part and Sum queries are independently decoded by different SA and CA layers (SA: Self-attention and CA: Cross-attention).}
\label{tab:decoder}
\scalebox{0.55}{
\begin{tabular}{c|cc cc|cccc}
\hline
Decoder design& \multicolumn{4}{c|}{Phrase Detection}& \multicolumn{4}{c}{Relationship Detection}\\ 
& \multicolumn{2}{c}{$k=1$}& \multicolumn{2}{c|}{$k=70$}& \multicolumn{2}{c}{$k=1$}& \multicolumn{2}{c}{$k=70$}\\
& R@50 & R@100 & R@50 & R@100 & R@50 & R@100 & R@50 & R@100  \\\hline 
Shared-stream&27.32&32.71&31.59&37.82&20.04&23.10&24.89&29.87 \\
Independent-stream&{\bf30.63}&{\bf 33.82}&{\bf 32.55}&{\bf 40.63}&{\bf 23.57}&{\bf 27.63}&{\bf 26.48}&{\bf 31.83} \\
\hline 
\end{tabular}}
\vspace{-3mm}
\end{table}

{\noindent\bf Results.} Table \ref{tab:hico-det1} shows the results of our method and the other state-of-the-art HOI methods on HICO-DET dataset. We see that most of the two-stage models have mAP around 20 (default, full) on the HICO-DET test set. The best two-stage model is DRG \cite{gao2020drg} which achieves 24.5 mAP. However, it is a complex model and requires three-stage training. In comparison, as end-to-end single stage models, our model and the contemporary HoiT \cite{zou2021_hoitrans} model are able to achieve 20+ mAP without using a dedicated object detector or extra pose or language information. Our model with composite queries has mAP of 23.9 and achieves the state-of-the-art performance for single-stage HOI.

\subsection{Ablation Study}
We report more components analysis of our proposed part-and-sum transformer model (PST) here.

{\noindent\bf Shared-stream vs independent-stream decoding}
Our part-and-sum transformer (PST) consists an independent-stream decoder for part queries and sum queries, i.e. part queries and sum queries are feed into variant self-attention layers, cross-attention layers and FFNs, and decoded independently. We compare this design with a shared-stream design, where part and sum queries are decoded by the same layers. The results are shown in Table \ref{tab:decoder} where the independent-stream design is shown to outperform the shared-branch design on both relationship detection and phrase detection tasks. We hypothesize that part queries and sum queries represent different aspects of a relationship, and it is better to decode these two kinds of queries independently.

{\noindent\bf Varying the number and dimension of the queries} 
We show the comparisons for the tensor-based query strategy vs the vector-based query strategy by varying their numbers and dimensions. Specifically, we use 500 tensor queries in PST, and each query contains three sub-vector queries of 256 dimension. For a fair comparison, we also employ 500 vector queries but of 256$\times$3 dimension. Note that increasing the dimension of vector-based queries to three times larger ($256\times3$) requires to three time larger dimension of image memory. To this end, we repeat the features of the encoder three times. By doing it, each query in both models can have the equal embedding dimension to represent each relationship. The comparison of results is shown in Table \ref{tab:tensor}. From it, we see that (1) increasing the number of vector queries from 500 to 1500 is not able to bring the clear benefit; (2) Increasing the dimension of each vector query does not provide more information; (3) Tensor based query outperforms variant vector based query, with a margin of 4.65\%/5.16\% increase on R@50/100 on Relationship detection when $k=1$; (4) Compared to the tensor based query, composite queries further improve the performance due to part-and-sum two-level learning.
 
\begin{table}[!ht]
\vspace{-2mm}
\centering
\setlength{\tabcolsep}{0.06cm}
\caption{\small Comparison of variant query designs in the PST.}
\label{tab:tensor}
\scalebox{0.60}{
\begin{tabular}{ccc|cc cc|cccc}
\hline
\multicolumn{3}{c|}{Query Design}& \multicolumn{4}{c|}{Phrase Detection}& \multicolumn{4}{c}{Relationship Detection}\\
Formulation &Number&Dimension & \multicolumn{2}{c}{$k=1$}& \multicolumn{2}{c|}{$k=70$}& \multicolumn{2}{c}{$k=1$}& \multicolumn{2}{c}{$k=70$}\\
&&& R@50 & R@100 & R@50 & R@100 & R@50 & R@100 & R@50 & R@100  \\\hline 
Vector&1500& 256 &26.39&29.68&29.41&32.68&18.63&21.14&20.26&23.64 \\
Vector&500&$256$&26.17&29.43&27.66&32.71&17.88&19.41&19.97&23.08 \\
Vector&500&$256\times3$&25.13&30.17&27.88&33.34&18.65&20.94&21.20&25.92 \\
Tensor&500&256&30.40&  {\bf 34.86}& 32.29& 37.68& 23.28&26.30& 25.46& 29.65 \\
Composite&500&256&{\bf30.63}& 33.82 &{\bf 32.55}&{\bf 40.63}&{\bf 23.57}&{\bf 27.63}&{\bf 26.48}&{\bf 31.83} \\
\hline 
\end{tabular}
}
\end{table}
{\noindent\bf Part-and-Sum design in the HOI task}
 The comparisons for the vector based Transformer (PST-Sum), tensor based Transformer (PST-Part), and part-and-sum Transformer (PST) on the HOI task are shown in Table \ref{tab:hoi}. PST outperforms PST-Part and PST-Sum, same as the performance comparison in the VRD task. Note that \cite{zou2021_hoitrans} can be regarded as a kind of PST-Sum model, but with different implementations such as different classification losses, and Transformer attention designs. 
 
  
\begin{table}[!htp]
\vspace{-2mm}
\centering
\caption{\small Results of variant PST decoders on HOI.}
\label{tab:hico-det2}
{
\scalebox{0.7}{
\begin{tabular}{c|ccc|ccc}
\hline
& \multicolumn{3}{c}{Default} & \multicolumn{3}{|c}{Known Object} \\
Method & Full$\uparrow$ & Rare$\uparrow$ & NonRare$\uparrow$ & Full$\uparrow$ & Rare$\uparrow$ & NonRare$\uparrow$ \\ \hline
PST-Sum  & 21.37 & 13.85 & 23.62 & 23.28 & 15.24 & 25.69 \\
PST-Part& 22.24 & 14.15 & 24.65 & 24.15 & 15.61 & 26.70 \\
PST& {\bf{23.93}} & \textbf{14.98} & \textbf{26.60} & \textbf{26.42} & \textbf{17.61} & \textbf{29.05} \\\hline 
\end{tabular}
}
\label{tab:hoi}}
\vspace{-2mm}
\end{table}

\vspace{-1mm}
\section{Conclusion}
\vspace{-1mm}
In this work, we have presented a Transformer-based detector, Part-and-Sum Transformers (PST), for visual relationship detection and human object interaction detection. PST maintains separate representations for the sum and parts while enhancing their interactions with composite queries. This design helps reduce instance ambiguity in structure data detection by learning rich intra-relationship and inter-relationship simultaneously. More qualitative results are provided in the supplementary document.
%
 
{\small
\bibliographystyle{ieee_fullname}
\bibliography{egbib}
}

\clearpage

\appendix

\noindent \textbf{\Large Appendix}

\section{Part-and-Sum Transformers with Composite Queries}
In this work, we focus on end-to-end structured data detection by Part-and-Sum Transformers for tasks like visual relationship detection and Human Object Interaction detection. We provide a more detailed discussion for the designs of \textbf{vanilla decoder, tensor-based decoder, and composite (part-and-sum) decoder}. These three alternatives differ in the form of queries and how attention is implemented. 

Figure 1(a) in the main submission gives an illustration; for an input image, we use a convolutional neural network (CNN) model to extract image features, which are fed into a standard~\cite{carion2020end}/ deformable~\cite{zhu2020deformable} transformer encoder. The encoder is composed of multiple self-attention layers to tokenize the visual features. After that, a transformer decoder takes the visual tokens together with a set of learnable queries as input to detect a composite set (visual relationships or human object interactions). We denote the tokenized features of the Transformer Encoder as $\boldsymbol{I}$, and the learnable queries as $\boldsymbol{Q}$; and the embedding of the outputs of a decoder as $\boldsymbol{E} = \text{Decoder}(\boldsymbol{Q},\boldsymbol{I})$. For embeddings $\boldsymbol{E}$, the structural prediction $\boldsymbol{O}$ is inferred by a prediction module, denoted as $\boldsymbol{O}=\text{Prediction}(\boldsymbol{E})$.

\subsection{Vanilla decoder with vector-based query}
Our vanilla Transformer decoder contains $M$ query embeddings, and each query is a vector, representing a relationship:
\begin{align}
      \boldsymbol{Q} &=\{\boldsymbol{q}_{1},...,\boldsymbol{q}_{M}\}, 
    \label{eq:vectq}
\end{align}
where $\boldsymbol{q}_{i}$ is a vector of a size $1\times D$, and the overall query $\boldsymbol{Q}$ is $M\times D$. The queries are feed into multiple decoder layers of a same design. Specifically, each decoder layer contains a Multi-head self-attention layer \cite{vaswani2017attention}, learning the cross-relationship context; and a multi-head cross-attention layer, to learn the representations by attending various image positions; and a feed forward network (FFN) to further embed each query. All query embeddings are feed into these three components one by one, and the last outputs are feed into the following decoder blocks. The decoding process in each decoder block is written as:
\begin{align}
\begin{split}
     f(\boldsymbol{Q}) & =\operatorname{SA}(\boldsymbol{q}_{1},...,\boldsymbol{q}_{M})\\
      \varphi(\boldsymbol{Q},\boldsymbol{I})   &=\operatorname{CA}([\boldsymbol{q}_{1},...,\boldsymbol{q}_{M}],\boldsymbol{I}),
    \end{split}
    \label{eq:vector}
\end{align}
where $f$ is Self-attention layer (SA), and $\varphi$ is the Cross-attention layer (CA). Note that in vanilla Transformer, each query represents a relationship, i.e. containing multiple components.
\subsection{Tensor-based decoder with tensor-based Query}
Unlike vector based query, tensor-based query represents a relationship by a tensor which contains multiple sub-queries to represent each part individually, such as Subject, Predicate and Object parts. The tensor based query can be written as:
\begin{align}
      \boldsymbol{Q}   &=\{\boldsymbol{Q}_{1},...,\boldsymbol{Q}_{M}\}
      =\{\{\boldsymbol{q}^{s}_{1},\boldsymbol{q}^{p}_{1},\boldsymbol{q}^{o}_{1}\},...,\{\boldsymbol{q}^{s}_{M},\boldsymbol{q}^{p}_{M},\boldsymbol{q}^{o}_{M}\}\},
    \label{eq:vectq2}
\end{align}
where each query $\boldsymbol{Q}_{i}$ includes three sub-queries $\boldsymbol{q}^{s}_{i},\boldsymbol{q}^{p}_{i},\boldsymbol{q}^{o}_{i}$ to represent subject, predicate, and object, respectively. By doing so, all parts are learnt individually, reducing the ambiguity in similarity computation in attention schemes. It is important for learning the relationships sharing the same subject or object entity. In decoding, attention layers handle all sub-queries, written as: 
\begin{align}
\begin{split}
     f(\boldsymbol{Q}) & =\operatorname{SA}(\boldsymbol{q}^{s}_{1},\boldsymbol{q}^{p}_{1},\boldsymbol{q}^{o}_{1},...,\boldsymbol{q}^{s}_{M},\boldsymbol{q}^{p}_{M},\boldsymbol{q}^{o}_{M})\\
      \varphi(\boldsymbol{Q},\boldsymbol{I})   &=\operatorname{CA}([\boldsymbol{q}_{1},...,\boldsymbol{q}_{M}],\boldsymbol{I}).
    \end{split}
    \label{eq:part2}
\end{align}
Vector query and tensor-based query are conceptually different, and the former learns each two-level/structure data as a whole/Sum, while the latter learns each two-level/structure data by part learning. Furthermore, self-attention layers are functionally different in these two designs: self-attention among all sum queries is to mine inter-relation context, while self-attention layer among part queries is to mine the context of entities, which indirectly benefits relationship learning.

\subsection{Composite (part-and-sum) decoder with composite Query}
Composite query models each relationship in a structural manner, and learns a relationship in both part and sum levels. Each composite query contains three part sub-queries for Subject, Predicate and Object entities, and one sum sub-query for a whole relationship. The composite query can be written as: 
\begin{align}
\begin{split}
      \boldsymbol{Q}   &=\{\boldsymbol{Q}_{1},...,\boldsymbol{Q}_{M}\} \\
      \boldsymbol{Q}_{i}&= \{\boldsymbol{q}^{s}_{i},\boldsymbol{q}^{p}_{i},\boldsymbol{q}^{o}_{i},\boldsymbol{q}^{\text{G}}_{i}\},  
    \label{eq:vectq3}
\end{split}
\end{align}
where $\{\boldsymbol{q}^{s}_{i},\boldsymbol{q}^{p}_{i},\boldsymbol{q}^{o}_{i}$ are part queries, and $\boldsymbol{q}^{\text{G}}_{i}$ is a sum query for relationship $i$. In the decoding, part query $\boldsymbol{Q}^{P}$ and sum query $\boldsymbol{Q}^{G}$ are separately decoded by different self-attention layers $f_{\text{Part}}$ and $f_{\text{Sum}}$, and cross-attention layers $\varphi_{\text{Part}}$ and $\varphi_{\text{Sum}}$, written as: 
\begin{align}
\begin{split}
    f_{\text{Part}}(\boldsymbol{Q}^{P}) & =\operatorname{SA}(\boldsymbol{Q}^{P}_{1},...,\boldsymbol{Q}^{P}_{M})\\
     \varphi_{\text{Part}}(\boldsymbol{Q}^{P},\boldsymbol{I}) & =\operatorname{CA}([\boldsymbol{Q}^{P}_{1},...,\boldsymbol{Q}^{P}_{M}],\boldsymbol{I}),
    \end{split}
    \label{eq:part3}
\end{align}

\begin{align}
\begin{split}
  f_{\text{Sum}}(\boldsymbol{Q}^{G}) & =\operatorname{SA}(q^{G}_{1},...,q^{G}_{M})\\
    \varphi_{\text{Sum}}(\boldsymbol{Q}^{G},\boldsymbol{I}) & =\operatorname{CA}([q^{G}_{1},...,q^{G}_{M}],\boldsymbol{I})
    \end{split}
    \label{eq:global}
\end{align}

\begin{figure*}[!th] 
    \includegraphics[width=\linewidth]{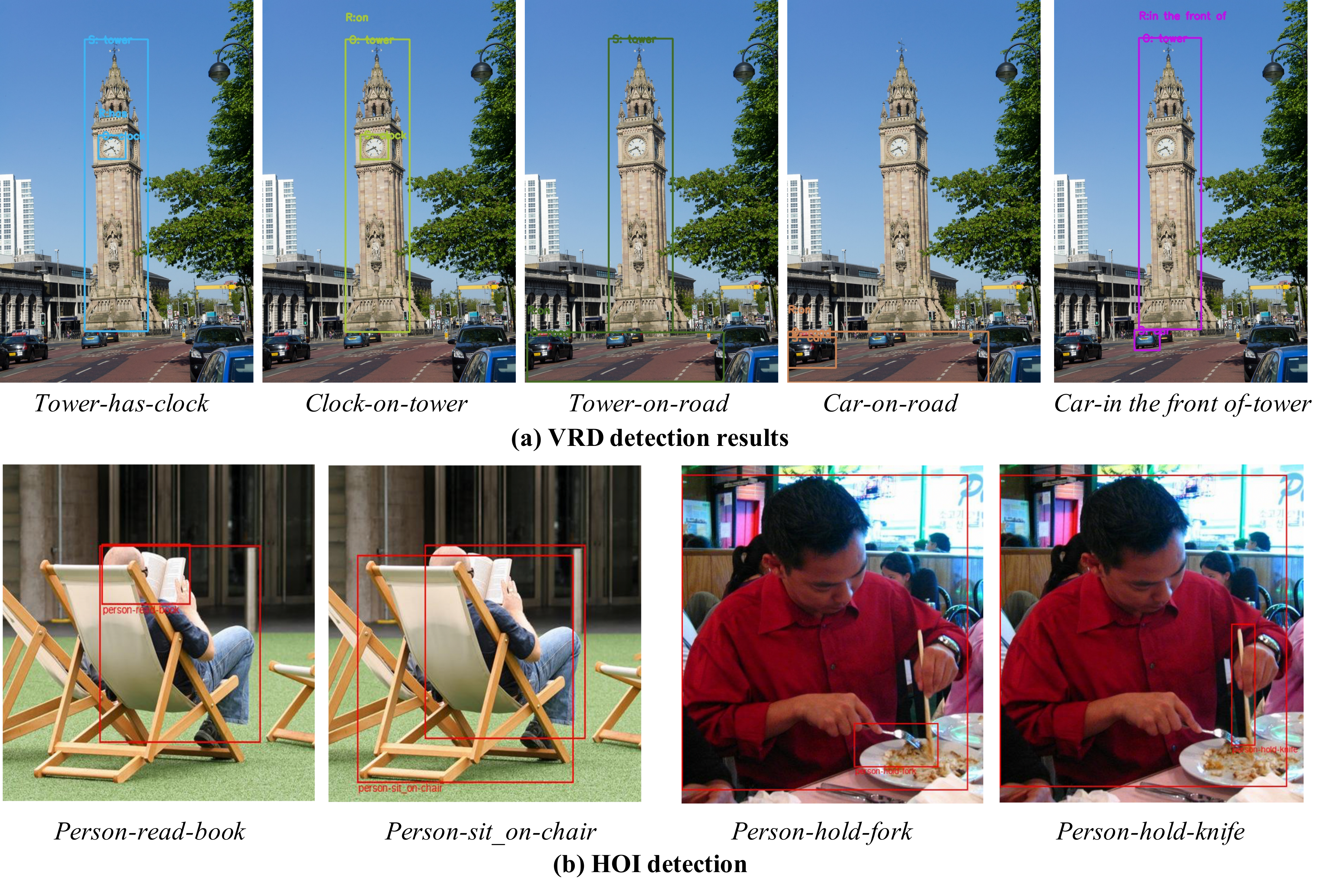}
    \centering
    \vspace{-5mm}
    \caption{\small Qualitative results on (a) VRD and (b) HOI by PST. Each sub-image shows one predicted relationship/interaction. ``R" refers to predicate; ``S" refers to subject; and ``O" refers to object.}
        \label{fig:det_vis}
\end{figure*}

{\noindent\bf{Factorized self-attention}}. To enhance part-based relationship learning, we designs a Factorized self-attention layer, which firstly conducts intra-relationship self-attention, and conducts inter-relationship self-attention. The intra-relationship self attention layer leverages the parts context to benefit relationship prediction, for example, subject query and object query are ``person" and ``horse" helps predict predicate ``Ride". The inter-relationship self-attention layer leverages the inter-relationship context, to enhance the holistic relationship prediction per image. For example, the existence of ``Person read book" helps infer the relationship ``Person sit", rather than ``Person run", which is particularly important for multiple interactions detection for same person entity. The Factorized self-attention is written as:
\begin{align}
\begin{split}
    f_{\text{Part}}(\boldsymbol{Q}^{P}) & =\operatorname{\text{FactorizedSA}}(\boldsymbol{Q}^{P}_{1},...,\boldsymbol{Q}^{P}_{M})\\
    & =\operatorname{Inter-relation SA}(\operatorname{Intra-relation SA}(\boldsymbol{Q}^{P})),\\
    \end{split}
    \label{eq:part4}
\end{align}
where Intra-relation self-attention and Inter-relation self-attention layers are written as:
\begin{align}
\begin{split}
    \operatorname{Intra-relation SA}(\boldsymbol{Q}^{P}_{i})&=\operatorname{SA}(q^{s}_{i},q^{p}_{i},q^{o}_{i})\\
    \operatorname{Inter-relation SA}(\boldsymbol{Q}^{P})&=\operatorname{SA}(\boldsymbol{Q}^{P}_{1},...,\boldsymbol{Q}^{P}_{M})\\
    \end{split}
    \label{eq:part5}
\end{align}
Note that the Factorized self-attention design also can be used for Tensor based query to enhance the inter part-query learning.
  \begin{figure*}[!th] 
     \includegraphics[width=0.8\linewidth]{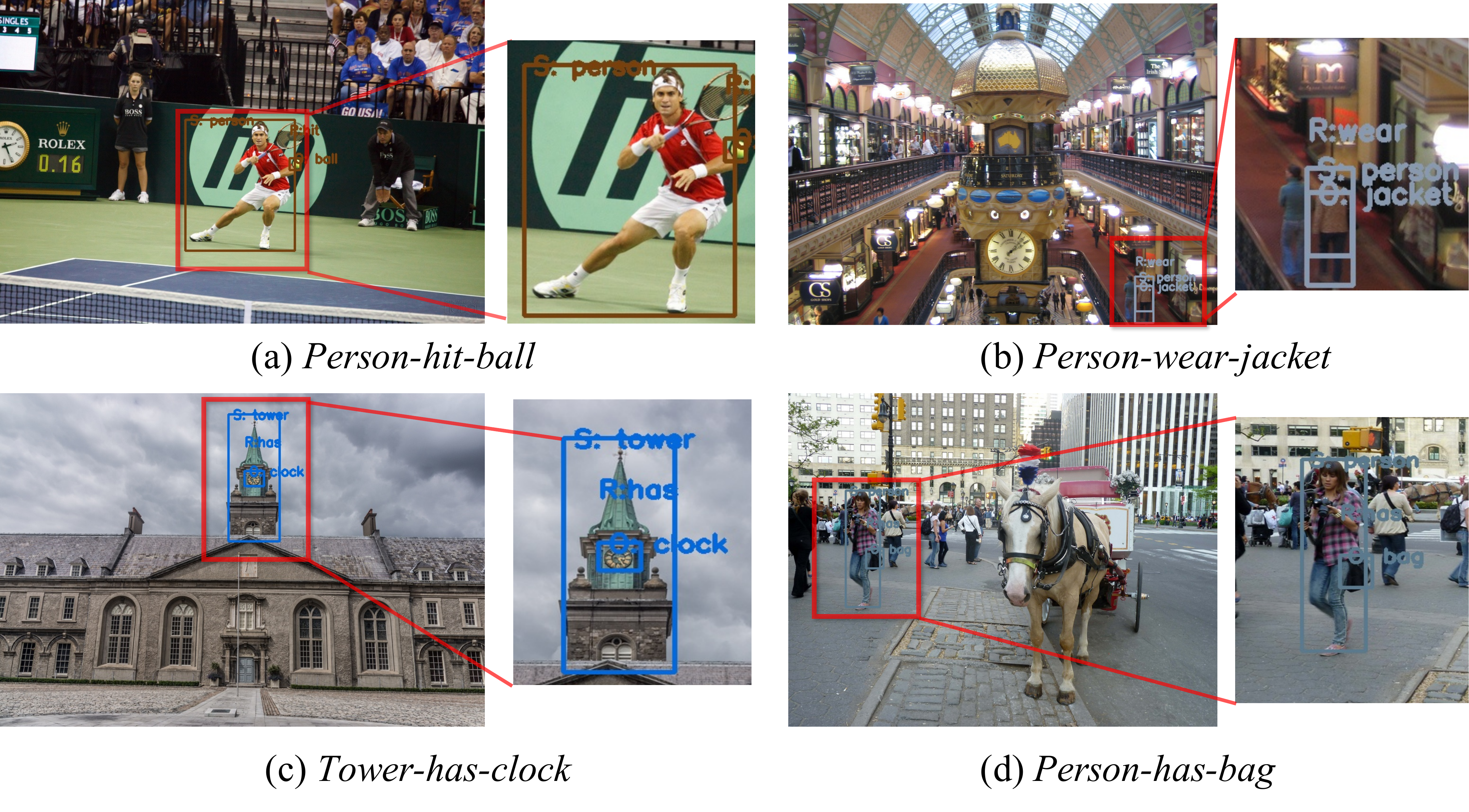}
     \centering
     \vspace{-2mm}
     \caption{\small Visualization of small-entity relationship detection. The exampled relationship predictions are (a) Person-hit-ball; (b) Person-wear-jacket; (c) Tower-has-clock; and (d) Person-has-bag. From it, PST is able to detect small subjects or objects in relationships and further detects the overall relationships properly. ``R" refers to predicate; ``S" refers to subject; and ``O" refers to object.}
         \label{fig:small_det_vis}
 \end{figure*}

 \begin{figure*}[!t] 
    \includegraphics[width=0.9\linewidth]{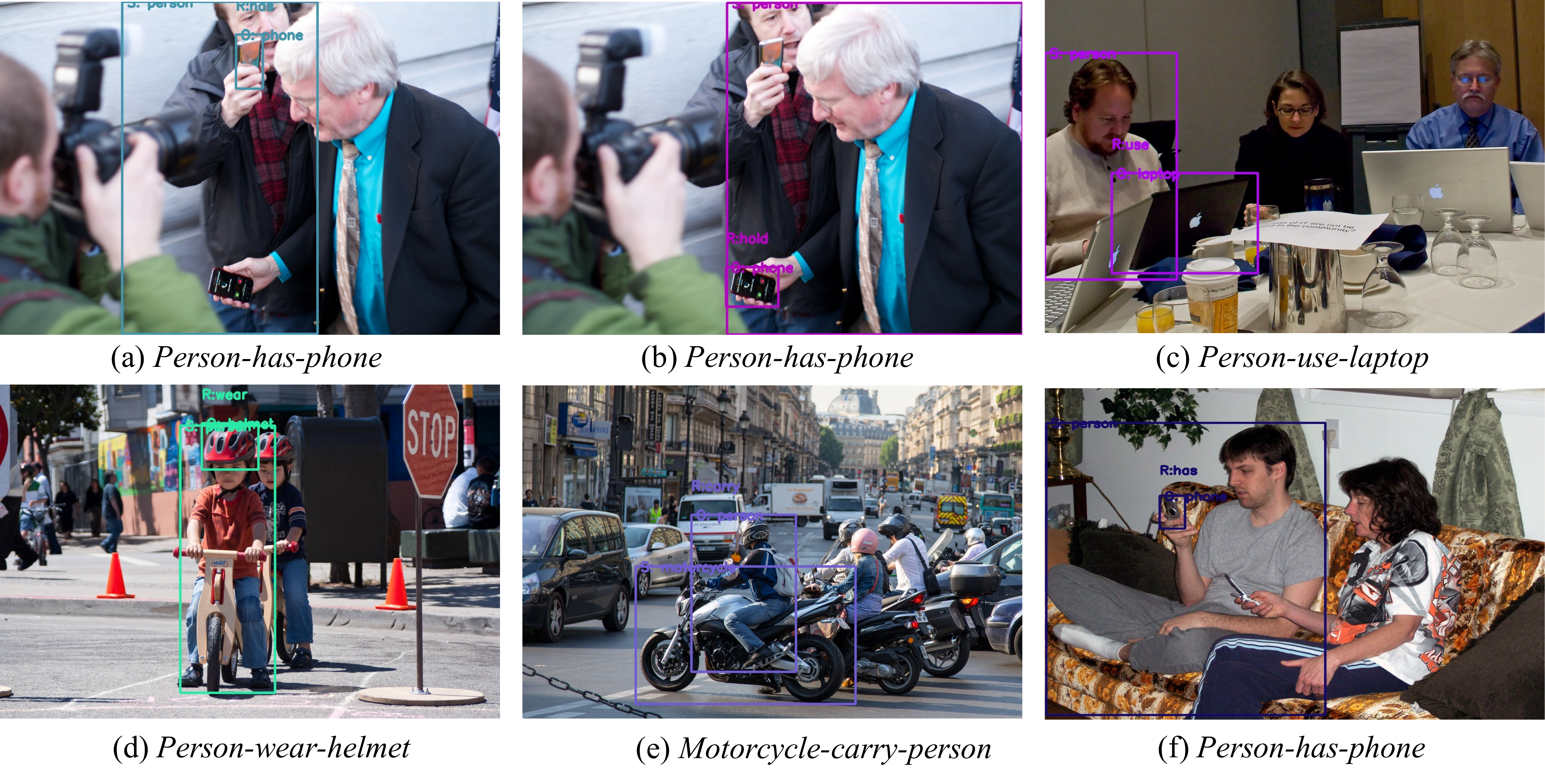}
    \centering
    \vspace{-2mm}
    \caption{\small Visualization of relationship detection with instance ambiguity. There exist multiple spatially close relationship instances of the same type. Specifically, there exists multiple same type and close relationships, for example, ``person-has-phone" in (a) and (b); ``person-use-laptop" in (c); ``Person-wear-helmet" in (d), ``Motorcycle-carry-person" in (e), and ``person-has-phone" in (f). ``R" refers to predicate; ``S" refers to subject; and ``O" refers to object.}
        \label{fig:multi_det_vis}
\end{figure*}

\begin{figure*}[!ht] 
    \includegraphics[width=1\linewidth]{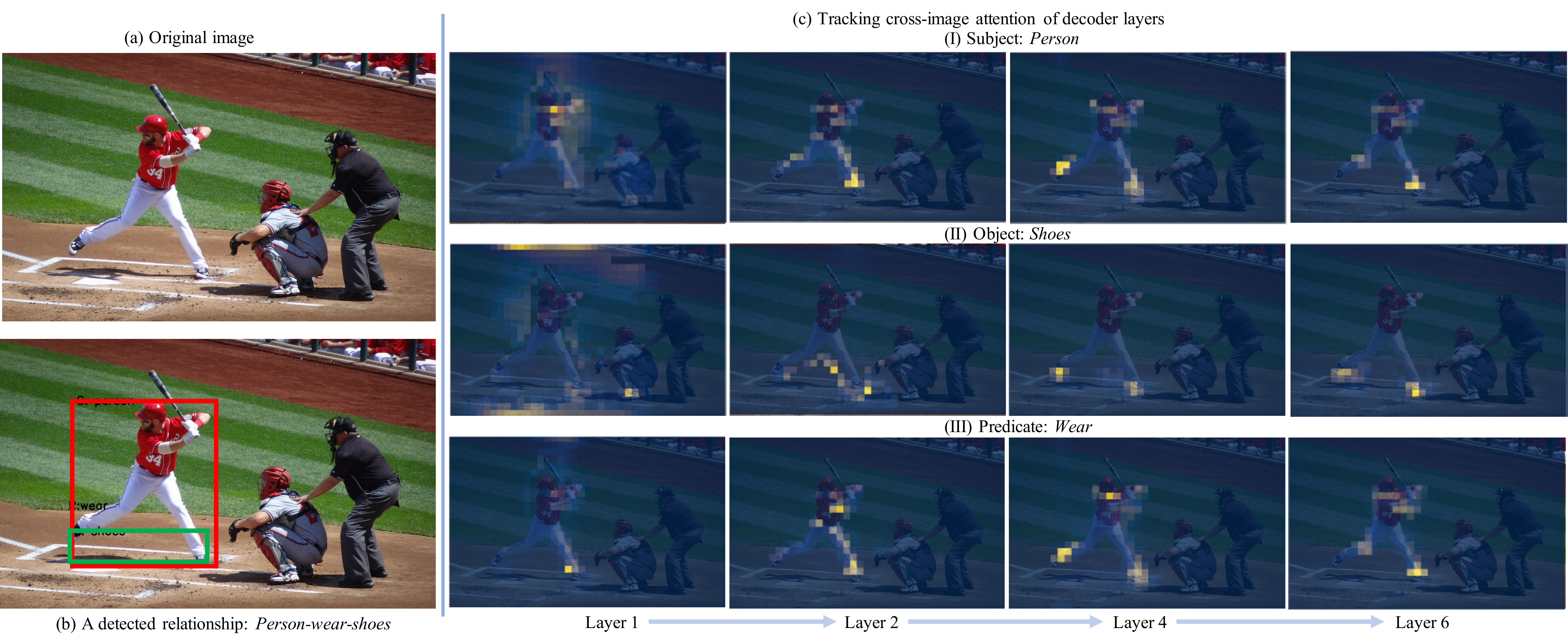}
    \centering
    \caption{\small Visualization of the attention maps of decoder layers in PST. (a) is the input image, (b) shows a detected relationship from one query, and (c) visualizes this relationship's attention maps in various decoder layers for subject (I), object (II) and predicate(III) query, respectively. Due to the space limit, we just show attention maps of four decoder layers.}
        \label{fig:attn_vis}
\end{figure*}
\section{Visualisations of VRD and HOI detection}
PST directly predicts all relationships in a set. Figure \ref{fig:det_vis} shows exampled relationship detection results and human object interaction detection results by PST in (a) and (b). Each sub-image visualizes one predicted relationship. It shows that there exist multiple relationships between one entity-pair. For example, in Figure \ref{fig:det_vis} (a), PST detects ``Tower-has-clock" and ``Clock-on-tower"; and ``Road-under-tower" and ``Tower-on-road".

\section{More Qualitative Illustrations}
In addition to the results shown in the main paper, we provide more qualitative results and analysis for visual relationship detection and human-object interaction detection by the proposed PST. 

\subsection{Small-entity relationship detection}
In the VRD task, relationships are composed of multiple types, some of which pose particular challenges, such as small-entity and spatial relationships.
We show some examples in Figure \ref{fig:small_det_vis} for small-entity relationship detection. PST is able to detect small subjects and objects well, such as ``ball" in (a), ``person" and ``jacket" in (b), ``clock" in (c) and ``bag" in (d). This happens because the part queries are able to mine the subject-predicate-object context, and the sum queries leverage the inter-relation context; the two types of contextual information provide effective information for detecting small entities in a relationship.

\subsection{Instance ambiguity in relationship detection}
Instance ambiguity in relationship detection causes a detection failure, where the predicted relationships wrongly associate subject and object instances, although the categories of relationships are predicted correctly. For instance, in Figure \ref{fig:multi_det_vis} (a) and (b), there exist two same type relationships ``Person-has-phone", and they are visually close. Relationship instance ambiguity makes it hard to associate each ``phone" instances with the surrounding ``Person" instances. This ambiguity is caused by that multiple relationship instances of the same relationship type are too close, and the visual clues for associating the subject-object instances are subtle. We examine PST in these challenging cases and show some examples in Figure \ref{fig:multi_det_vis}. It shows that PST is able to associate subject-object instances correctly in this hard situation, thanks to the effective intra-relation and inter-relation attention for context modeling. 
 \begin{figure}[!th] 
    \includegraphics[width=0.9\linewidth]{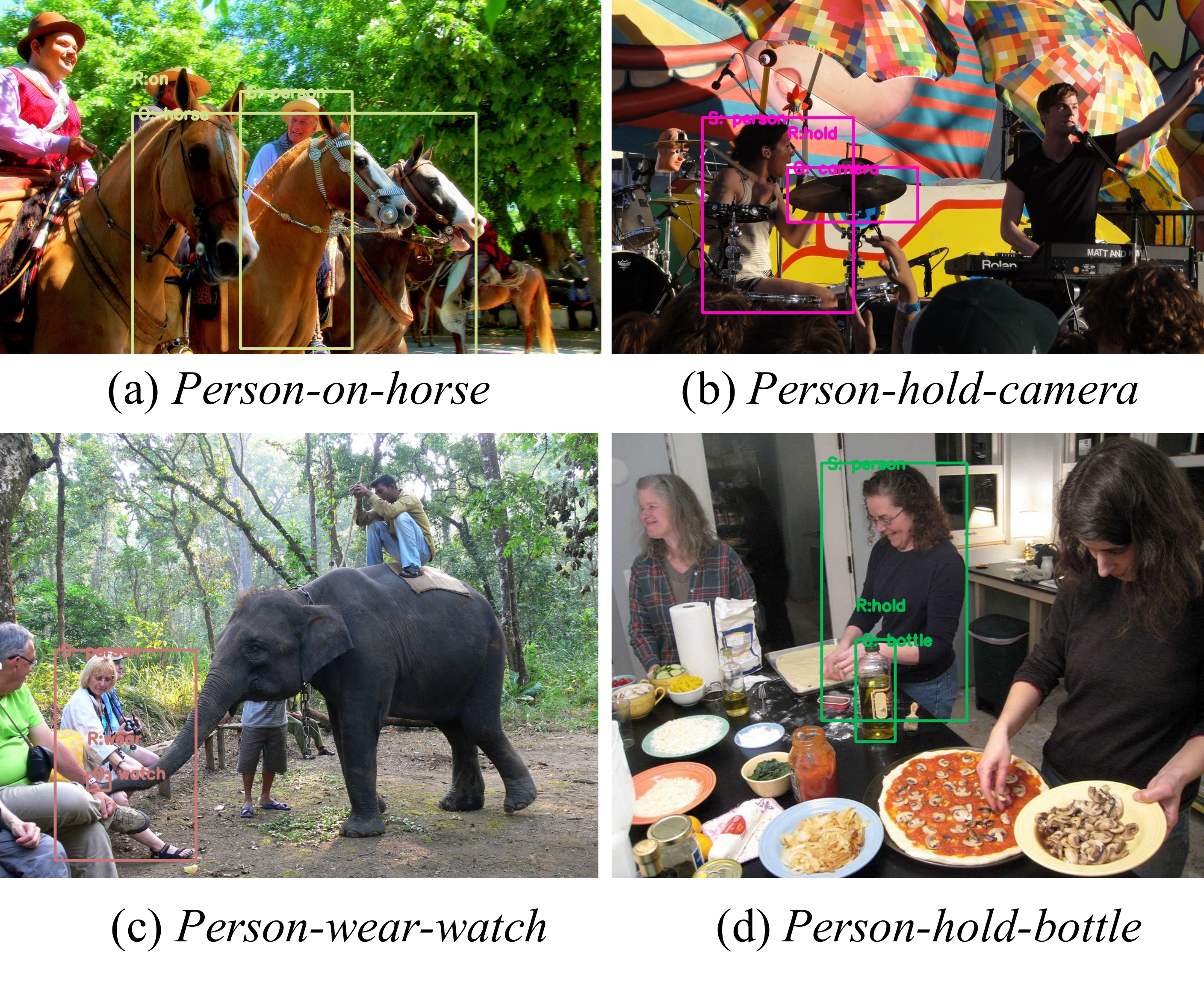}
    \centering
    \vspace{-4mm}
    \caption{\small Visualization of a few failure cases by PST. There are four main relation detection types: (a) Inaccurate entity detection caused by crowed instances; (b) Wrong object detection; (c) Wrong association between subject and object instances; (d) Wrong predicate classification. 
    ``R" refers to predicate; ``S" refers to subject; and ``O" refers to object.
    }
        \label{fig:failure}
        \vspace{-2mm}
\end{figure}
\subsection{Composite attention visualisation}
To better understand how the model works and what input information it uses to perform relationship detection, we visualize the cross-attention maps of the decoder layers of PST, since cross-attention measures the correlation between the query embedding and image feature tokens. 
Given a test image, we extract cross-attention maps from all decoder layers for the subject, object and predicate query embedding individually. We visualize the attention maps of one query in Figure \ref{fig:attn_vis} (c), and include results of more queries in the supplementary material. In Figure~\ref{fig:attn_vis}, the relationship query embedding is decoded to {\em``person-wear-shoes"} semantically, and according to the attention maps, we can see that (1) the transformer decoder {\em incrementally} focuses on the ``person" and ``shoes" area in the image, to infer the subject and object entities; (2) the predicate (``{\em Wear}") is mostly predicted from the union area of the subject and object, which suggests that the attention scheme is capable of automatically modeling the subject-object context for predicate detection.

\subsection{Failure cases by PST}
We visualize the typical errors of relationship detection by PST in Figure \ref{fig:failure}. There are four typical errors: (1) PST localizes entities inaccurately, when the entity instances are crowed. For instance, in Figure  \ref{fig:failure} (a), there are multiple horses close to each other, and PST localizes multiple horses as one Object entity of a relationship ``Person-on-horse". (2) Object detection mistakes cause the failure in relationship detection, such as wrong entity ``camera" detected in Figure \ref{fig:failure} (b). (3) Relationship instance ambiguity challenges PST. For instance, in Figure \ref{fig:failure} (c), the watch is associated with a wrong person instance which is very close to the right person instance. (d) Predicate is wrongly predicted, for instance, PST classifies the relationship between ``Person" and ``bottle" as ``hold".

\end{document}